\documentclass[a4paper,fleqn]{cas-dc}

\usepackage[authoryear,longnamesfirst]{natbib}

\def\tsc#1{\csdef{#1}{\textsc{\lowercase{#1}}\xspace}}
\tsc{WGM}
\tsc{QE}

\begin{document}
\let\WriteBookmarks\relax
\def\floatpagepagefraction{1}
\def\textpagefraction{.001}

\shorttitle{A Study of the Quality of Wikidata}    

\shortauthors{Shenoy, Ilievski, Garijo, Schwabe, Szekely}  

\title [mode = title]{A Study of the Quality of Wikidata}  

\newcommand{\dg}[2][inline]{\color{red} [DG:]: #2\color{black}}
\newcommand{\filip}[2][inline]{\color{blue} [FI]: #2\color{black}}
\newcommand{\kartik}[2][inline]{\color{orange} [KS]: #2\color{black}}



%



\author[1]{Kartik Shenoy}


\ead{kshenoy@isi.edu}

\credit{Credit}

\affiliation[1]{organization={Information Sciences Institute, University of Southern California},
            citysep={}, 
            country={USA}}
            
\affiliation[2]{organization={Ontology Engineering Group, Universidad Politécnica de Madrid},
            citysep={}, 
            country={Spain}}
        
\affiliation[3]{organization={Dept. of Informatics, Pontificia Universidade Cat\'olica Rio de Janeiro},
            citysep={}, 
            country={Brazil}}

\author[1]{Filip Ilievski}
 \cormark[1]

\ead{ilievski@isi.edu}


\author[2]{Daniel Garijo}
\ead{daniel.garijo@upm.es}

\author[3]{Daniel Schwabe}
\ead{dschwabe@inf.puc-rio.br}

\author[1]{Pedro Szekely}
\ead{pszekely@isi.edu}

\credit{Credit}


\cortext[1]{Corresponding author}



\begin{abstract}
Wikidata has been increasingly adopted by many communities for a wide variety of applications, which demand high-quality knowledge to deliver successful results. In this paper, we develop a framework to detect and analyze low-quality statements in Wikidata by shedding light on the current practices exercised by the community. We explore three indicators of data quality in Wikidata, based on: 1) community consensus on the currently recorded knowledge, assuming that statements that have been removed and not added back are implicitly agreed to be of low quality; 2) statements that have been deprecated; and 3) constraint violations in the data. We combine these indicators to detect low-quality statements, revealing challenges with duplicate entities, missing triples, violated type rules, and taxonomic distinctions. Our findings complement ongoing efforts by the Wikidata community to improve data quality, aiming to make it easier for users and editors to find and correct mistakes.
\end{abstract}




\maketitle



\section{Introduction}
\label{sec:intro}


\setcounter{footnote}{0} 

Historically, Wikipedia is 
the best known knowledge base relying on the ``wisdom of the crowd'' \citep{surowiecki2004woc} to ensure its quality; setting an example for other popular websites such as Quora\footnote{\url{https://quora.com/}} and Stack Exchange.\footnote{\url{https://stackexchange.com/}}
Wikidata~\citep{vrandevcic2014wikidata} has been created in a similar manner - editing it is fairly straightforward. Consequently, Wikidata today is a joint creation of tens of thousands of human and bot contributors~\citep{piscopo2018onto}. The result is a rich set of factual statements that describe claims about entities and events in the real world. New information is entered everyday, resulting in very high growth rates and immediate description of popular world events.\footnote{In this work, we use the terms \textit{statement}, \textit{claim}, and \textit{fact} interchangably.}


Wikidata aims to allow ``plurality of 
facts''~\citep{mollersurvey}, and hence it is important that these facts are described with high-quality statements. We have little understanding of the quality of the knowledge contained in Wikidata.
Relatively simple validators can spot syntactic errors, allowing for automatic detection (`flagging' or editing) of syntactically anomalous statements~\citep{beek2014lod}. Yet, capturing and correcting semantic information is more challenging. While existing work has proposed an extensive set of quality notions~\citep{piscopo2019we}, and started to apply statement validation to Wikidata~\citep{thornton2019using,piscopo2018onto}, 
to our knowledge, no past work has comprehensively applied indicators to measure quality of statements in Wikidata as a whole, and provided a vision for improving its quality in the future. 
%
 



In this paper, we develop a framework to detect and analyze low-quality statements in Wikidata
by shedding light on the current practices exercised by the community. In addition, we propose to enhance the quality of Wikidata by automatically flagging potential problematic statements for editors. Our work makes the following contributions:


\begin{enumerate}
    \item We define three indicators that measure well-understood notions of quality of Wikidata statements, based on: 1) the statement revision history of Wikidata; 2) deprecation of statements; and 3) violations of property constraints defined by the community. 
    \item We develop an efficient framework that flags potential errors integrating these three indicators of quality. Namely, the community-based indicators find low-quality statements which have been deleted or deprecated throughout the history of Wikidata (since its inception in 2014), while the constraint-based indicator reveals outliers with high constraint violation ratios.  
    \item We apply our framework to analyze the quality of the entire Wikidata.\footnote{There are 1,149,471,184 statements in the Wikidata dump of December 2020.} We report findings on key aspects of quality that affect users and editors, such as low-quality type statements, taxonomical modeling errors, duplicated nodes, and missing statements. 
    \item We propose recommended actions to interactively support high-quality contributions in the future, as well as to retroactively fix existing issues. By doing so, we complement ongoing efforts by the Wikidata community to improve data quality based on games and suggestions, aiming to make it easier to prevent, find, and correct mistakes.
\end{enumerate}




Our quality indicators evaluate the degree of community consensus on what is acceptable, thus connecting to existing metrics of Wikidata quality, like accuracy, consistency, and veracity~\citep{piscopo2019we}.
By analyzing statements which have been removed, we reflect on the accuracy of the data.
By formulating and analyzing semantic rules (constraints) that statements must satisfy, we provide insights into the well-formedness and consistency of the data.
The analysis of the deprecated statements addresses the veracity of claims, by indicating that there was once consensus about their veracity, but this is no longer the case. 


We make our code\footnote{\url{https://github.com/usc-isi-i2/wd-quality}} \citep{kartik_shenoy_2021_5119983} and materials\footnote{\url{https://w3id.org/wd_quality}} available to facilitate further work on analyzing quality of Wikidata statements.
The rest of the paper is structured as follows. Section~\ref{sec:method} introduces the three indicators, their formalization, and combination into a joint framework. All of our findings with their supporting analyses are described in Section~\ref{sec:findings}. Recommended actions can be found in Section~\ref{sec:discussion}. We relate to prior work on Wikidata quality in Section~\ref{sec:relatedwork}. The paper concludes in Section~\ref{sec:conclusion}. 


\section{Framework}
\label{sec:method}



We seek to measure semantic 
quality aspects of Wikidata. We devise a framework for detecting low-quality statements in Wikidata, which combines three indicators of quality, based on: 1) community updates; 2) deprecated statements; and 3) property constraints. In this section, we describe each of the quality indicators and we provide details on their formalization into an integrated framework that can analyze the quality of Wikidata. We formalize low quality with $q=0$.

Throughout this section, we use $S$ to refer to a set of statements. A \emph{statement} $(s,p,o,Q) \in S$ refers to the union of an edge subject $s$, predicate $p$, object $o$, and qualifier set $Q$. Qualifier sets contain property-value pairs $(q_{p_i}, q_{v_i})\in Q$ that further describe the tuple $(s,p,o)$, (e.g., with the date or the source of assertion). Such statements are common building blocks of modern hyperrelational Knowledge Graphs (KGs), like Wikidata or YAGO \citep{tanon2020yago}. 




\subsection{Quality Indicators}
\label{ssec:indicators}
\textbf{Community-based indicator}
We define a community-based indicator of KG quality by considering that the KG statements that have been permanently deleted by the community (i.e., statements deleted at a time point $t_i$ and not restored in time points $t_{j}, j > i$) are of low quality. 
Following the idea of ``wisdom of the crowd''~\citep{surowiecki2004woc}, we assume that community-based KGs, like Wikidata, are self-correcting over time, i.e., its contributors detect low-quality statements, and either delete or replace them.

However, the set of removed tuples by itself is neither necessary nor sufficient to indicate incorrect statements. A statement might be simply updated with a semantically equivalent one. Object values may be reassigned from one property or class to another, which might be considered more appropriate to express the relationship between the subject and the object. Literals may be updated with a new value that may or may not be semantically different than the original one. The latter case often corresponds to the adoption of new naming conventions, e.g., replacing the name ``Pamela C Rasmussen'' with ``Pamela C. Rasmussen''. To address these issues, we consider the low-quality ($q=0$) statements of a dump $d$ at a time $t_i$ to be a union of: 1) the removed statements which were not updated ($R(d_{t_i})$), and 2) the removed statements which were updated with a significantly different value ($U(d_{t_i})$).
Formally, $S_c(q=0, d_{t_i})=R(d_{t_i}) \cup U(d_{t_i})$.




\textbf{Deprecation-based indicator}
Wikidata has a `soft' alternative to deletions: deprecating statements to indicate consensus about the end of their validity. A statement is marked as \emph{deprecated} in two cases: 1) if it has been superseded by another statement, or 2) if it is now known to be wrong, but was once thought correct.\footnote{\url{https://www.wikidata.org/wiki/Help:Deprecation}}
For example, the community agreed that Pluto ceased to be a planet since 13th September, 2006 and hence the claim stating that fact has been deprecated. 

Deprecated statements ($D$) are valuable for studying the evolution of Wikidata and the agreement about its statements. However, they are undesired when using Wikidata in applications that require up-to-date information, like entity linking and question answering. Thus, we consider all deprecated statements of a dump $d$ at a time $t_i$ to be indicators of low quality, formally: $S_d(q=0, d_{t_i})=D(d_{t_i})$.

\textbf{Constraints-based indicator}
The Wikidata community has defined property constraints, i.e., rules that specify how properties should be used.
\footnote{\url{https://www.wikidata.org/wiki/Help:Property\_constraints\_portal}} 
Each property in Wikidata specifies the constraint types that apply to it.
Statements expressed with that property can then either conform to the constraint or violate it. We denote the set of all violations in a Wikidata dump $d$ at a time $t_i$ with $V(d_{t_i})$. 
Constraints are split in three groups: mandatory, suggested, and normal (i.e., constraints which are neither mandatory, nor suggested). Each constraint type is further specified per property, by stating additional \textit{elements}: property-dependent classes, exceptions, and property paths. 



\begin{figure*}[t]
    \centering
    \includegraphics[width=\textwidth]{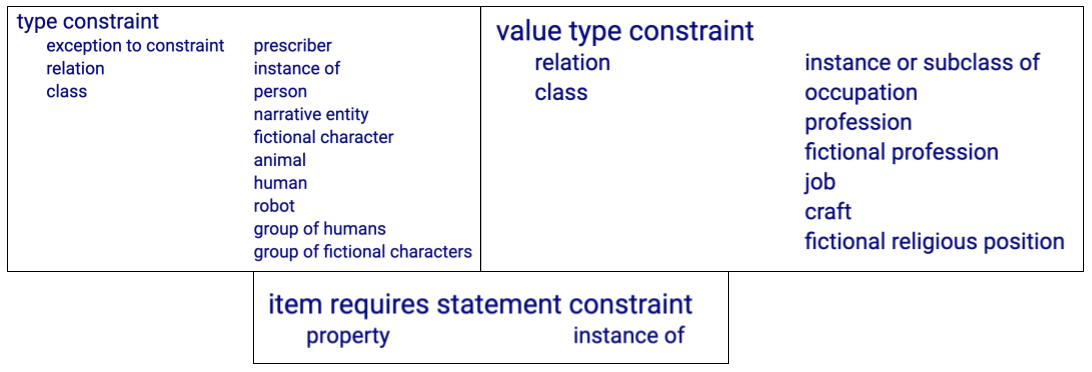}
    \caption{Example constraints for the property \texttt{occupation (P106)}: \textit{type} (top-left), \textit{value type} (top-right), and \textit{item requires statement} (bottom). The \textit{type} constraint specifies that subjects that have an occupation have to be instances of one of the eight allowed classes, unless the subject is \texttt{prescriber}. The \textit{value type} constraint dictates that objects of \texttt{occupation} statements have to be either instances or subclasses of one of the six possible classes shown. The \textit{item requires statement} constraint specifies that items which have an \texttt{occupation} value must also have an \texttt{instance-of} statement. All depicted constraints have a normal 
    status.}
    \label{fig:constraints}
\end{figure*}

At present, Wikidata defines 30 types of property constraints. Constraints vary in nature, and range from format validation (e.g., correct dates or naming conventions) to ensuring a consistent usage of a property (e.g., making sure that symmetric properties are used in both directions).
We provide examples for three key constraint types in Figure~\ref{fig:constraints}: type constraint, value type constraint, and item-requires-statement constraint. 
The Wikidata type and value type constraints 
indicate that the domain of a property (or range, respectively) has to conform to one of the listed classes, but specify them further with exceptions and property paths.
The item-requires-statement constraint dictates that a Wikidata item with one property should also specify another one. Constraints may also specify \textit{exceptions}.
In Figure~\ref{fig:constraints}, the \textit{type} constraint indicates that subjects that have an occupation have to be instances of one of the eight allowed classes, unless the subject is \texttt{prescriber} (``person legally empowered to write medical prescriptions''),\footnote{\url{https://www.wikidata.org/wiki/Q99393050}} whereas the \textit{value type} constraint dictates that objects of \texttt{occupation} statements have to be either instances or subclasses of one of the six possible classes shown.
The \textit{item requires statement} constraint specifies that items which have an \texttt{occupation} value must also have an \texttt{instance-of} statement. All constraints presented in this figure have a normal status.

The constraint-based indicator considers violations of property constraints in their corresponding statements to be low-quality statements.
We denote the set of statements that violate a constraint with $V$. The set of low-quality ($q=0$) statements according to this indicator is: $S_v(q=0, d_{t_i})=V(d_{t_i})$.

\subsection{Experimental Setup}


While the three indicators of quality have different foci, each of them identifies a set of low-quality statements, denoted by $S_c$, $S_d$, and $S_v$ in the previous section. In the rest of the paper, we analyze the low-quality statements identified by each indicator. 
We inspect deprecated and permanently deleted statements in Wikidata, we assess what constraints are violated, and we compare the violations with the deletions. 
In our experiments, we employ the Knowledge Graph ToolKit (KGTK)~\citep{ilievski2020kgtk}, which supports flexible and scalable imports of Wikidata, and supports efficient manipulation of large hyperrelational KGs, which is essential for the analysis carried out by our quality framework.




\textbf{Community-based indicator:}
We collected a dataset of Wikidata statements
that have been permanently removed (i.e., removed and not added again) since the first available dump of Wikidata in October, 2014. The dumps of Wikidata are released weekly. We generated this dataset by downloading all available weekly JSON Wikidata dumps from the Internet Archive,\footnote{\url{https://archive.org/search.php?query=wikidata}} resulting in 311 dumps;\footnote{Approximately two years of dumps were missing from Internet Archive, but we were able to retrieve them with the help of contributors from the Wikidata community.} converting them to the KGTK format; and extracting statements that had been removed between each pair of successive dumps $(d_{t_i}, d_{t_j})$, where $t_i<t_j$. We also checked whether statements that have been removed before $t_i$ were present in the more recent of the two dumps, $d_{t_j}$. Formally:

$A(d_{t_i},d_{t_j}) = d_{t_j} \setminus d_{t_i}$, with $t_i<t_j$

$R(d_{t_i},d_{t_j}) = d_{t_i} \setminus d_{t_j}$, with $t_i<t_j$

$Tr(d_{t_j}) = (Tr(d_{t_i})\setminus A(d_{t_i},d_{t_j})) \cup R(d_{t_i},d_{t_j})$, with $t_i<t_j$

Here, $A$ and $R$ represent the added and deleted statements between $d_{t_i} $ and $d_{t_j}$, respectively. The operator $\setminus$ represents a difference between two sets, $\cup$ is the union, and the total removed statements for the 0-th dump is $Tr(d_{t_0}) = \emptyset$




After obtaining the full set of removed statements, we analyzed how many of the nodes had been redirected to new nodes (i.e., duplicate removal), and computed the distribution of classes and properties being removed. For literals, we investigated whether a value had been entirely removed or updated by computing the similarity between the removed value and the new one. We analyzed the similarity for each literal type separately. For strings, we measured Levenshtein distance between the removed and the updated text. For dates, we measured the time distance between the removed and the updated date. For quantities, we computed the difference in magnitude between the removed and the new quantity. We consider deleted statements with no update and deleted statements with a notable update to be of low quality (cf. Section~\ref{ssec:indicators}).



\textbf{Deprecation-based indicator:}
We consider all deprecated statements to be of low quality.
Wikidata indicates deprecation through the \texttt{rank} qualifier of a statement.
We retrieved all statements with a deprecated rank value in the early Jan, 2021 version of Wikidata (the last dump we collected), and we explored their distribution in terms of entities and properties.

\begin{figure}
\small
\begin{verbatim}


kgtk query 
-i statements instance_of subclass_of_star \
--match statements: (subject)-[id {label:property}]->(object), \
        instance_of: (subject)-[]->(class), \  
        subclass_of_star: (class)-[]->(parent)' \
--where 'parent in expected_parents or subject in exceptions' \
--return 'distinct id, subject, property, object' \

\end{verbatim}
 \caption{Example of a query template for the \textit{type} constraint. The set of allowed parent classes for the subject are defined in \texttt{expected\_parents}, whereas \texttt{exceptions} is the set of subjects for which this constraint is not required. If the subject of a statement is an instance of a class in \texttt{expected\_parents}, or any of its subclasses, then the constraint is satisfied for that statement. The constraint is also satisfied if the subject belongs to the set of \texttt{exceptions} defined by the property constraint. Notably, for some properties, the \texttt{instance of} relation is replaced with a \texttt{subclass of} relation.}
    \label{fig:query}
\end{figure}

\textbf{Constraint-based indicator:}
We consider statements that violate constraints to be of low quality, $q=0$.
We prioritized constraints that are common in Semantic Web research and cover a sufficient number of properties (e.g., type and value type). 

Wikidata has pages with constraint violation reports,\footnote{\url{https://www.wikidata.org/wiki/Wikidata:Database\_reports/Constraint\_violations}} which are calculated with an ad-hoc extension of Wikibase.\footnote{\url{https://gerrit.wikimedia.org/r/plugins/gitiles/mediawiki/extensions/WikibaseQualityConstraints}} However, it is unclear whether these reports are updated regularly.
Given the size of Wikidata, validating its constraints with the Shape Constraint Language (SHACL) or the Shape Expressions Language
(ShEx) is computationally prohibitive~\citep{boneva2019shape}. Moreover, it is unclear whether these languages can encode exceptions and allowed values in property constraints, and, to the best of our knowledge, there is no available implementation of SHACL/ShEx constraint validators for Wikidata. 
For this reason, we encoded each constraint type as a KGTK query template. 
Each template is instantiated once per property
, allowing their efficient validation in parallel. Constraint violations for a property are computed in a two-step manner: we first obtain the set of statements that satisfy the constraint for a property, and then we subtract this set from the overall number of statements for that property.
We omit constraints defined on external identifier properties, as our aim is to capture semantic and modeling errors in Wikidata.

An example query template is shown in Figure~\ref{fig:query}. The query inspects whether the subjects for a property are instances of a class that is allowed by the constraint, or any of its subclasses. If this is the case, or the subject is listed as an exception to the constraint, then the constraint is satisfied for this statement. Notably, for some properties, the \texttt{instance of} relation is replaced with a \texttt{subclass of} relation. A full example query for one property can be seen in Annex  
\ref{ann:queryExample}.



\begin{table*}[t!]
    \centering
    \caption{Statistics of the constraints: type (Q21503250), value type (Q21510865), item requires statement (Q21503247), inverse (Q21510855), and symmetric (Q21510862). We show the number of properties with  (M)andatory, (N)ormal and (S)uggested constraints, and the corresponding number of statements. For the \textit{item requires statement} constraint type $all \leq M+N+S$, because properties have multiple constraints with a potentially different status.}
    \label{tab:time}
    \begin{tabular}{l |r r r r | r | r r r r} \toprule
        & \multicolumn{4}{c}{\#properties} & \multicolumn{1}{c}{\#statements} & \multicolumn{4}{c}{validation time (in sec.)}\\
        constraint type& all & M & N & S & \multicolumn{1}{c}{all} & min & max & mean & median \\ \midrule
        type & 1,456 & 165 & 1,280 & 11 & 513,424,170 & 4.95 & 5231.15 & 366.16 & 174.78 \\
        value type & 897 & 106 & 786 & 5 & 182,087,480 & 11.41 & 5323.18 & 352.08 & 144.15 \\
        item requires statement & 527 & 78 & 418 & 97 & 302,642,146 & 1.89 & 2199.57 & 133.51 & 58.6 \\
        inverse & 110 & 6 & 100 & 4 & 9,440,925 & 8.68 & 646.22 & 100.69 & 54.79 \\
        symmetric & 38 & 5 & 30 & 3 & 7,145,197 & 9.72 & 527.33 & 118.44 & 68.67 \\
        
\bottomrule
    \end{tabular}
\end{table*}

\textbf{Combination of indicators:} Each quality indicator produces a set of statements. We compute the overlap between the deleted statements and the constraint violations as follows.
We added all deleted statements to the Wikidata version where we computed the violations, and calculated the number of violations without and with the total removed statements: $V$ and $V_{del}$, respectively. The difference between these two yields the number of violations that were fixed by the removal of the statements ($V_{fixed}$). Formally:

$V = V(d_{t_i})$

$V_{del} = V(d_{t_i} \cup Tr(d_{t_i}))$

$V_{fixed} = V_{del} \setminus V$

\section{Findings}
\label{sec:findings}

Our framework indicators result in: 1) a dataset of 76.5M removed statements, describing 26.2M distinct subjects \citep{garijo_daniel_2021_4686805}; 
2) a dataset of 10M deprecated statements \citep{shenoy_kartik_2021_5120117}; 
and 3) a set of correct statements and constraint violations \citep{shenoy_kartik_2021_5121276}, according to the constraint types specified in Table~\ref{tab:time}. This table shows that most of the property constraints have a normal status, and that the median time to validate a property constraint over Wikidata ranges between 55 and 175 seconds for the five constraints. This demonstrates the feasibility of our approach to validate Wikidata constraints at scale. 

In this section, we highlight the main findings of our analysis by shedding light into complex issues related to KG quality, such as node redundancy, naming conventions, taxonomic distinctions, completeness, accuracy of constraints, and type consistency. We also explore whether constraint violations are getting corrected over time, thereby improving the overall quality of Wikidata. 
Specifically, we study the following eight research questions:
\begin{enumerate}
    \item Are entities being deduplicated?
    \item Can the community distinguish classes from instances?
    \item Are naming conventions needed?
    \item Are property types and value types respected?
    \item Can we detect missing triples?
    \item Are constraints correct and complete?
    \item What statements get deprecated?
    \item Are constraint violations getting fixed?
\end{enumerate}

For each of these questions: 1) we motivate its relevance and impact on Wikidata; 2) we present our findings about its current state; and 3) we provide an in-depth analysis and representative examples. Based on these findings, we provide recommendations about improving the state-of-the-art quality of Wikidata in Section \ref{sec:discussion}.






\subsection{Are Entities being Deduplicated?}


Entity linking and deduplication are complex open research challenges in many KGs.  
Redirects are a common mechanism to deduplicate nodes, and are applied when a user recognizes that two nodes describe the same subject, e.g., \texttt{Category:1911 in Morocco} redirects from  \texttt{Q18511155} to \texttt{Q9404406}.\footnote{\url{https://www.wikidata.org/wiki/Help:Redirects}} Our analysis reveals over 2 million redirected nodes, which affect over 20 million statements (26\% of all removed statements). The relatively high number of redirects reflects Wikidata's dynamic nature and the community pursuit for a high-quality, well-integrated graph. It is not known how many duplicate entities currently remain in Wikidata.

21.3 million statements (27.8\% of the removed statements) have either a redirected subject or a redirected object.
We inspected the property containing the largest number of redirected items, \texttt{instance of (P31)}, to understand what type of nodes have been redirected. Table \ref{tab:rclass} (top) shows the five classes with the highest number of redirected instances, which include well-populated classes in Wikidata like human, scholarly article, and gene. In addition, a portion of the \texttt{instance of (P31)} redirects are due to classes that themselves have been redirected. Table \ref{tab:rclass} (bottom) shows the five redirected classes with a highest number of member instances, which include encyclopedic article, village of Poland, and rotating variable star.

\begin{table*}[t!]
    \centering
    \small
    \caption{Distribution of classes in redirected P31 statements. We show 5 classes with the highest number of redirected instances, and 5 classes that have been redirected themselves. The counts and the percentages represent numbers of affected statements. The percentages are relative to total redirected statements, not total statements.} 
    \label{tab:rclass}
    \begin{tabular}{l l r } \toprule
        \multicolumn{3}{c}{\bf Classes of redirected instances} \\
        \midrule
        Q4167836 &
        Wikimedia category &
        526,207 (21.38\%) \\
        Q5 &
        human &
        222,809 (9.05\%) \\
        Q4167410 &
        Wikimedia disambiguation page &
        108,583 (4.41\%)\\
        Q13442814 &
        scholarly article &
        101,156 (4.11\%) \\
        Q7187 &
        gene &
        88,231 (3.59\%) \\ \midrule
        \multicolumn{3}{c}{\bf Redirected classes} \\ \midrule
        Q17329259 &
        encyclopedic article &
        301,359 (12.25\%) \\
        Q4423781 &
        dictionary entry &
        53,671 (2.18\%) \\
        Q17143521 & village of Poland &
        51,581 (2.09\%) \\
        Q15917122 &
        rotating variable star &
        50,642 (2.06\%)\\
        Q20900710 &
        painting & 
        23,482 (0.99\%)\\

        \bottomrule

    \end{tabular}
\end{table*}






\begin{table*}[t!]
    \centering
    \caption{Community updates of instance-of (P31) and subclass-of (P279).}
    \label{tab:taxonomy}
    \begin{tabular}{l l r c} \toprule
        \bf before & \bf after & \bf count & \bf example \\ \midrule
        P31 & P31 & 2.85M & (Hardenstein Castle,P31,geographical feature) \\ 
        & & & $\rightarrow$ (Hardenstein Castle,P31,ruins) \\\hline
        & P279 & 44k & (laboratory centrifuge,P31,laboratory equipment)\\
        & & & $\rightarrow$ (laboratory centrifuge,P279,laboratory equipment) \\\hline
        & both & 106k & (mystic,P31,person) \\
        & & & $\rightarrow$ (mystic,P31,non-professional work activity)\\
        & & & $\rightarrow$(mystic,P279,religious) \\\hline
        & none & 703k & (Clubland Smashed,P31,album)$\rightarrow$\textit{none}
 \\\hline
        P279 & P31 & 444k & (Chemical Markup Language,P279,markup language)\\
        & & & $\rightarrow$ (Chemical Markup Language,P31,markup language) \\\hline
        & P279 & 33k & (girder bridge,P279,bridge by structural type) \\
        & & & $\rightarrow$ (girder bridge,P279,bridge) \\\hline
        & both & 421k & (barn,P279,building) \\
        & & & $\rightarrow$(barn,P31,type of farm house)\\
        & & & $\rightarrow$(barn,P279,agricultural structure)\\
        & & & $\rightarrow$(barn,P279,appendage) \\\hline
        & none & 36.5k & (Categoria:Plantilles d'informació de videojocs,P279,\\
        & & & Category:Wikimedia templates) $\rightarrow$\textit{none}\\\bottomrule
    \end{tabular}

\end{table*}

\subsection{Can the Community Distinguish Classes from Instances?}

When adding new instances to Wikidata, contributors 
must specify descriptive values for
the taxonomy relations of \texttt{instance of (P31)} and \texttt{subclass of (P279)}. 
Wikidata's fairly wide ontology (containing millions of classes) 
and the prior evidence on the difficulty of distinguishing between taxonomic relations in Wikidata~\citep{piscopo2018onto}, 
raise the question: can the community distinguish classes from instances? Our analysis of removed statements with object properties reveals nearly half a million cases where one of the taxonomic relations has been changed to the other, which point to the fact that the community struggles to decide whether to use instance-of (P31) or subclass-of (P279) to model inheritance in Wikidata.\footnote{\url{https://www.wikidata.org/wiki/Wikidata:WikiProject\_Ontology/Problems}} 

Drilling down, we see that in 44 thousand cases, the \texttt{instance of} statement was replaced with a \texttt{subclass of} statement. In the case of former P279 edges, the number of taxonomic switches is notably larger: nearly half (444k out of 935k) P279 edges were replaced by a P31 edge only. Illustrative examples in Table~\ref{tab:taxonomy} indicate that these switches often happen in cases where it is not trivial to distinguish between the two taxonomic relations. For example, the community struggles to specify the membership of laboratory centrifuge as laboratory equipment - a former \texttt{instance of} relation has been replaced with a \texttt{subclass of} one. Conversely, the Chemical Markup Language used to be specified as a \texttt{subclass of} a markup language, but this has been corrected into an \texttt{instance of} relation. In both cases, the updated relation seems more intuitive, which, in line with the ``wisdom of the crowd'' assumption, would indicate that switches between the two relations largely reflect fixes of prior modeling errors.

\begin{figure*}[t]
    \centering
    \includegraphics[width=0.75\textwidth]{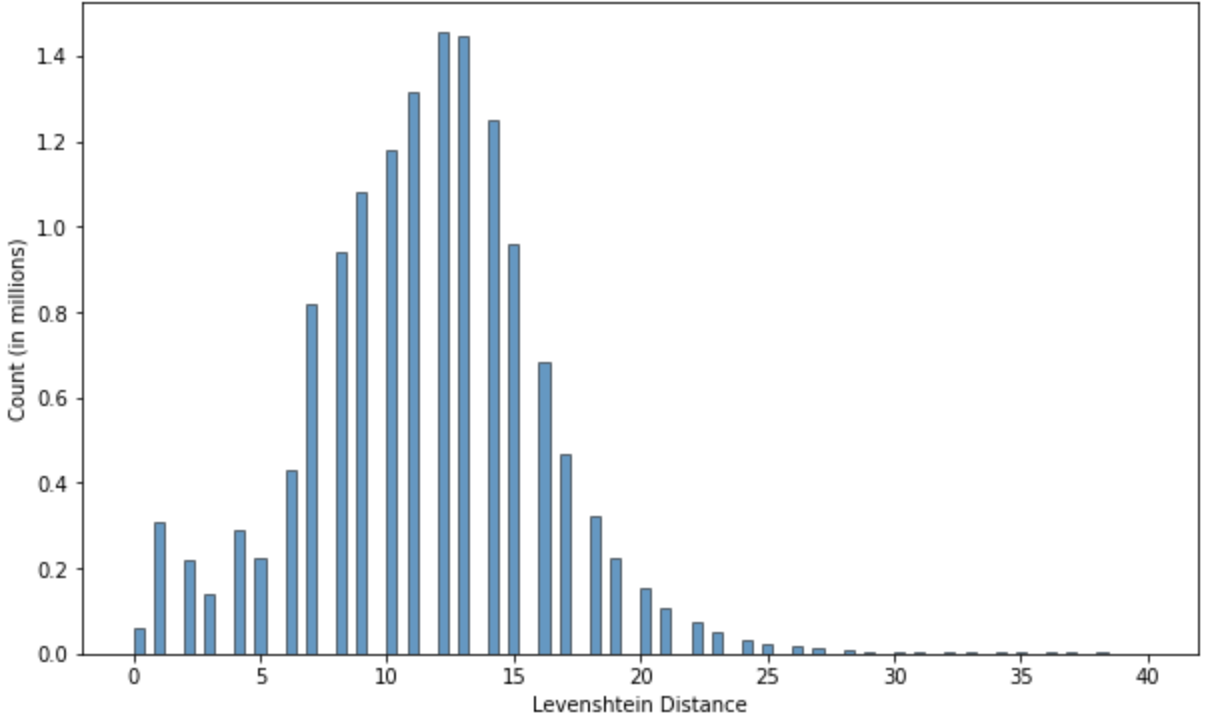}
    \caption{Distribution of the Levenshtein distance between old and new string values. The x-axis shows the Levenshtein distance, while the y-axis shows number of statements in each bucket, in terms of millions.}
    \label{fig:lev}
\end{figure*}

\subsection{Are Naming Conventions Needed?}

To our knowledge, Wikidata does not prescribe how to encode strings, though there are guidelines for dates.\footnote{\url{https://www.wikidata.org/wiki/Help:Dates}} 
We performed an analysis to investigate the proportion of updates for both strings and dates, in order to study current practices, and possible oscillations between different semantically equivalent values. Our analysis reveals that the community has already performed millions of updates between semantically (nearly) equivalent forms of literals.

In particular, we observe that in the majority of cases (61.5\% of all removed dates), the date was replaced with a semantically equivalent date with a different surface form. An example is the year 1964, modified from ``000000001964-00-00T00:00:00Z/9'' to ``1964-00-00T00:00:00Z/9''. When it comes to removed string statements, we observe that 46\% of them (14 million) have been replaced with new values. The distribution of the Levenshtein distances between the old and the new string values is shown in Figure \ref{fig:lev}. 
We observe that strings with low Levenshtein distances are typically stylistic updates, e.g., from ``Pamela C Rasmussen'' to ``Pamela C. Rasmussen''. Among the strings with a medium Levenshtein distance (of 10), we see updates which are meant as specifications and can also be interpreted as mere stylistic adaptations, such as the update of ``Hiroshima EAST BLD'' to ``Hiroshima East Building''. The strings with a large distance (of 20) are generally different from the original strings, such as the update of ``Meredith Boyle Metzger'' to ``Susan Michaelis''.

\begin{table*}[t!]
    \centering
    \small
    \caption{Correct (constraint-satisfying) and incorrect (constraint-violating) statements for the five constraint types analyzed in this paper: type (Q21503250), value type (Q21510865), item requires statement (Q21503247), inverse (Q21510855), and symmetric (Q21510862). The violation ratio (VR) is the percentage of incorrect statements in the total set of statements in a given category. We separate the statistics among (M)andatory, (N)ormal and (S)uggested constraints.}
    \label{tab:violations}
    \begin{tabular}{l | r r r | r r r | r r r} \toprule
        & \multicolumn{3}{c}{mandatory} & \multicolumn{3}{c}{normal} & \multicolumn{3}{c}{suggested}\\
        constraint type & correct & incorrect & VR\% & correct & incorrect & VR\% & correct & incorrect & VR\% \\ \midrule
        type & 44.99M & 37.67k & 0.08 & 464.71M & 3.58M & 0.76 & 85.03k & 21.65k & 20.29 \\
        value type & 11.44M & 5.38k & 0.03 & 169.47M & 1.11M & 0.65 & 46.15k & 512 & 1.09 \\
        I.R.S. & 3.98M & 767 & 0.02 & 272.71M & 2.25M & 0.82 & 25.73M & 2.24M & 8.01 \\
        inverse & 6.56k & 133 & 1.99 & 7.13M & 0.21M & 2.79 & 2M & 95.35k & 4.55 \\
        symmetric & 7.43k & 42 & 0.56 & 6.23M & 78.88k & 1.25 & 0.77M & 54.22k & 6.55 \\
        
        
        \bottomrule
    \end{tabular}
\end{table*}

\begin{figure*}
    \centering
    \includegraphics[width=0.74\textwidth]{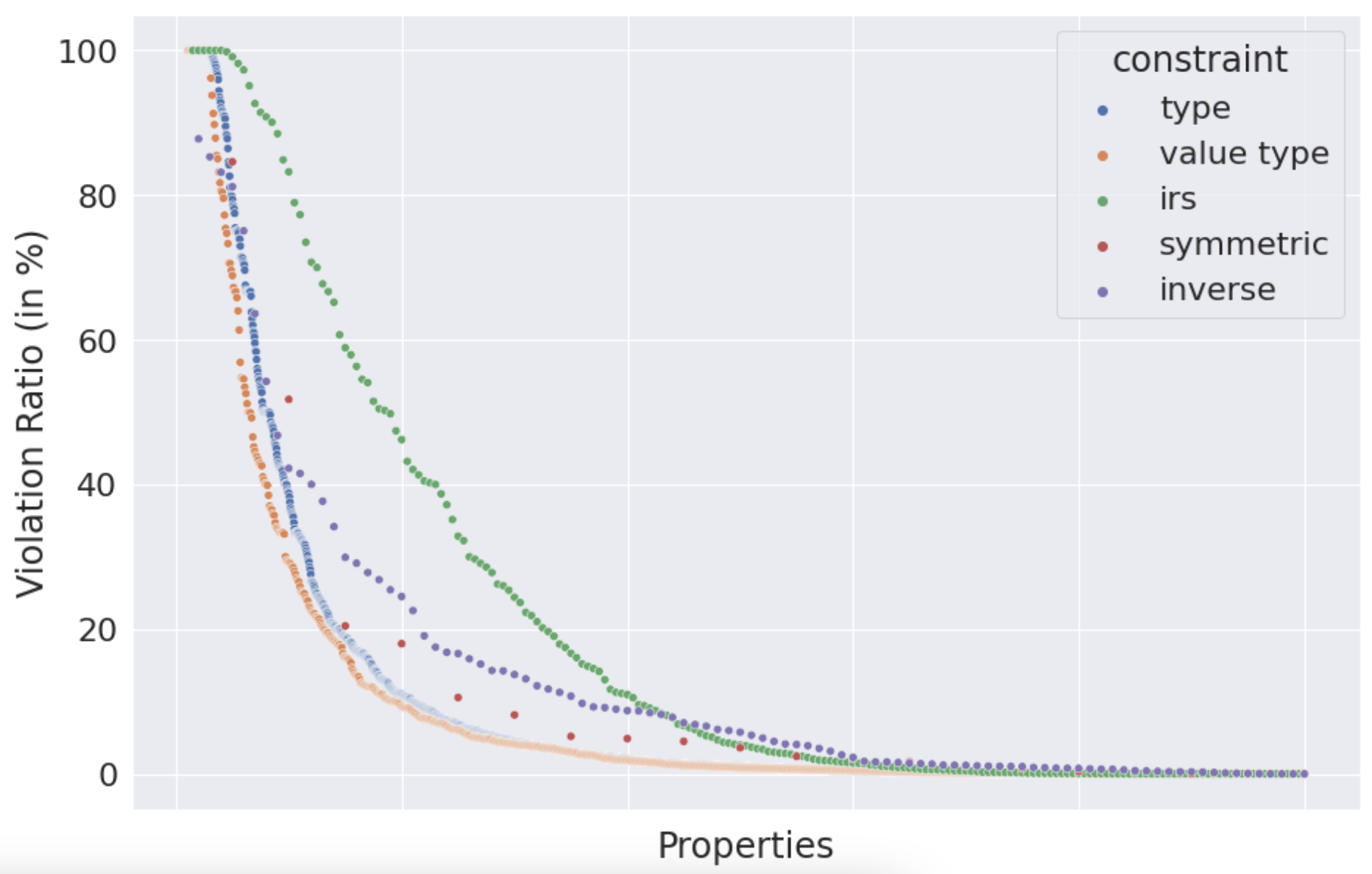}
    \caption{Distribution of Violation Ratios (VRs) for each of the five constraints. Each dot corresponds to a single property. Properties are shown in a descending order according to their VRs. The number of points varies according to constraint usage.} 
    \label{fig:vrs}
\end{figure*}

\subsection{Are Property Types and Value Types Respected?}



Type and value type constraints are similar to the domain and range constraints in Semantic Web languages like OWL, and are covered in resources like YAGO~\citep{tanon2020yago} and VerbNet~\citep{schuler2005verbnet}. Many properties in Wikidata have associated type and value type constraints, as shown in Table~\ref{tab:time}. Have these constraints been respected by the data? We observe that only a small portion of the mandatory constraints, and a much larger portion of the suggested constraints, violate the set constraints. While the violations are largely concentrated around a small set of properties and could in theory be fixed, it is unclear whether this is desired, as the suggested status implies that they might not need to be strictly enforced.

As shown by the violation ratios in Table~\ref{tab:violations} (rows 1 and 2), only a small portion of the mandatory type and value type constraints are violated (0.08\% and 0.03\%, respectively). The proportion of violations is larger for normal constraints, which represent the majority (0.76\% and 0.65\%, respectively). The violation ratio is the highest for the suggested constraints, where as many as 20\% of the statements were found to violate type constraints. This might be expected, as the status \textit{suggested} implies less strict semantics than mandatory constraints. This analysis entails that fixing the current type and value type violations would require nearly 44 thousand edits for the mandatory constraints, and 4.7 million edits for the normal and suggested constraints. Figure~\ref{fig:vrs} shows a Zipfian distribution of the violation ratios for the properties that have type and value type constraint, i.e., most violations are concentrated around a few properties.

\begin{table*}[t!]
    \centering
    \caption{Top-3 constraint violations for each constraint type. The violation ratio (VR\%) is the percentage of incorrect statements in the total set of statements for a given property.}
    \label{tab:vexamples}
    \begin{tabular}{l l l | r r} \toprule
        constraint & property & label & VR\% & \#statements \\ \midrule
        type & P8138 & located in the statistical territorial entity & 100 & 461 \\
        & P5051 & towards & 100 & 64 \\
        & P2303 & exception to constraint & 100 & 39 \\ \hline
        value type & P5008 & on focus list of Wikimedia project & 100 & 331,026 \\
        & P6104 & maintained by WikiProject & 100 & 9,764 \\
        & P7374 & educational stage & 100 & 32 \\ \hline 
        i.r.s. & P1111 & votes received & 100 & 46,327 \\
        & P2302 & property constraint & 100 & 42,211 \\
        & P3063 & gestation period & 100 & 549 \\ \hline
        
        inverse & P1605 & has natural reservoir & 94.03 & 201 \\
        & P3448 & stepparent & 87.97 & 4,849 \\
        & P926 & postsynaptic connection & 85.71 & 7 \\ \hline
        symmetric & P5188 & Sandbox-Lexeme & 100 & 2 \\
        & P1706 & together with & 92.85 & 56 \\
        & P521 & scheduled service destination & 75.67 & 1,718 \\
\bottomrule
    \end{tabular}
\end{table*}

\subsection{Can we Detect Missing Triples?}

It is well known that broad-coverage KGs are inherently incomplete~\citep{dong2014knowledge}. This incompleteness can be partially addressed through the property constraints: item-requires-statement (IRS), inverse, and symmetric. These constraints point to a missing triple for the same entity, a missing triple with an inverse property, and with a symmetric property, respectively. For example, IRS dictates that entities that have an \texttt{occupation} property must also have a statement with the \texttt{instance of} property.
We investigate to which extent these constraints have been followed by the statements in Wikidata. As shown in Table \ref{tab:violations}, the mandatory constraints for these constraint types reveal nearly a thousand violations, which may indicate missing triples. The situation 
worsens for normal and suggested constraints, whose enforcement would lead to millions of potentially missing triples.
While fixing symmetric and inverse constraints is programmatically trivial, it is unclear whether this is always desired, as the constraint violations may be caused by an incorrect original statement rather than a missing one. For example, if a \texttt{spouse} link exists from entity $E_1$ to $E_2$, but not from $E_2$ to $E_1$, it is impossible to infer automatically whether $E_1$ and $E_2$ are spouses (in which case a link from $E_2$ to $E_1$ is missing) or not (in which case the link from $E_1$ to $E_2$ should be removed).

Table~\ref{tab:violations} (rows 3-5) 
illustrates how mandatory IRS and inverse constraints are largely followed 
(with only 0.02\% and 1.9\% violations, respectively). As expected, the violation ratios are larger for normal, and largest for suggested constraints, peaking at 8\% for the IRS suggested constraints. Table~\ref{tab:vexamples} shows examples for properties with highest violation ratios. For instance, the property \texttt{votes received (P1111)} requires other properties like \texttt{office contested (P541)} to be present, which is violated in all 46k cases where it appears. The inverse property for the properties \texttt{has natural reservoir (P1605)}  and \texttt{step\-parent (P3448)} is missing in nearly all cases, resulting in five thousand violations. The most commonly violated symmetric properties include \texttt{Sandbox-Lexeme (P5188)}, \texttt{together with (P1706)}, and \texttt{scheduled service destination (P521)}, resulting in around 1,500 violations in total.



\subsection{Are Constraints Correct and Complete?}
\label{ssec:complete}

If the constraints are to be used as a driving force to improve the quality of Wikidata, it is important that they are correct and complete. As shown in Table~\ref{tab:violations}, the majority of the constraints fit the data, which can be seen as an indicator that the constraints are of good quality. Yet, we note that across all constraint types, a small portion of the constraints yields a large portion of violations. 

The head of the distribution in Figure~\ref{fig:vrs} reveals properties whose constraint definitions are outdated. 
Table~\ref{tab:vexamples} lists those property constraints with large (nearly 100\%) violation ratios, which may point to discrepancies between the constraints and the underlying data. 
For example, \texttt{towards (P5051)} expects subjects to be instances of \texttt{transport stop (Q548662)}, which is violated for all its 64 instances. 28 of these instances have a type \texttt{vein (Q9609)} (e.g., external jugular vein (Q2512768)), and use the \texttt{towards} property to indicate the direction blood flow of a vein in the human body (e.g., subclavian vein  is oriented towards the brachiocephalic vein).
In this case, rather than fixing each statement with a constraint violation manually, one could generalize the constraint, 
i.e., enhance the type constraint for the \texttt{towards} property to allow for instances of \texttt{vein}. 

\begin{table*}[t!]
    \centering
    \caption{Top classes (left) and properties (right) in the deprecated statements.}
    \label{tab:deprecated}
    \begin{tabular}{l r | l r} \toprule
        \bf Class & \bf Count & \bf Property & \bf Count \\ \midrule
        infrared source (Q67206691) & 2,546,256 & instance of (P31) & 3,303,204\\
        star (Q523) & 352,194 & proper motion (P2215) & 2,236,125 \\
        near-IR source (Q67206785) & 60,055 & parallax (P2214) & 2,159,860\\
        astronomical radio source (Q1931185) & 43,618 & radial velocity (P2216) & 816,191\\
        galaxy (Q318) & 35,768 & distance from Earth (P2583) & 461,113 \\\bottomrule
    \end{tabular}

\end{table*}

\begin{table*}[!t]
    \centering
    \caption{Violations within the removed statements for each type of constraint.}
    \label{tab:overlap}
    \begin{tabular}{l | r | r | r} \toprule
        \bf constraint & \bf mandatory & \bf normal & \bf suggestion \\ \midrule
        \bf type &
        763k/2.31M (33.04\%) & 5.3M/34.87M (15.21\%) & 920/2.29k (40.12\%) \\
        \bf value type &
        25.4k/211k (12.03\%) & 198k/8.99M (22.06\%) & 235/397 (59.19\%) \\
        \bf IRS & 4.67k/1.28M (0.36\%) & 192k/4.85M (3.97\%) & 190k/6.01M (3.17\%) \\
        \bf inverse & 37/345 (10.72\%) & 177k/534k (33.13\%) & 11.7k/160k (7.27\%) \\
        \bf symmetric & 19/307 (6.19\%) & 7.52M/10.85M (69.37\%) & 5.05k/37.5k (13.47\%) \\ \bottomrule
    \end{tabular}
\end{table*}

\subsection{What Statements Get Deprecated?}

We investigate whether deprecated statements, as a soft alternative to deletions, reveal different behavior compared to removed statements. Among the 10 million statements with deprecated rank in Wikidata, we observe that many belong to the domain of Astronomy. This indicates that the decision between removing and deprecating a statement largely depends on the community and the domain.

Specifically, we found 10,040,256 deprecated statements. The top-5 properties in deprecated statements are shown in Table~\ref{tab:deprecated}. We observe that all frequently deprecated properties (e.g., proper motion) belong to the domain of Astronomy, and that large portion of the overall deprecations (around 90\%) is expressed with these first five properties. In addition, we observe that the deprecated \texttt{instance of} statements describe membership of celestial objects, like infrared source, star, and galaxy.

\subsection{Are Constraint Violations Getting Fixed?}

Our analysis reveals that Wikidata has millions of deleted statements and constraint violations. Do these two sets overlap? We observe that many of the removed statements violated a constraint, i.e., many of the removals coincide with former violations, thereby improving the quality of Wikidata over time.

Specifically, out of the 2.31 million removed statements for which a mandatory type constraint is defined, a third violated that constraint (Table~\ref{tab:overlap}).
Most of the former violations correspond to normal and suggested constraints. 
Overall, we observe that the removed statements fixed millions of constraint violations, including 6 million type violations and 7.5 million symmetric violations.

Notably, constraints could have been fixed or violated through the addition (instead of removal) of statements, which we are not considering in our work and, as such, it is a limitation of our current analysis.





\section{Recommendations} 
\label{sec:discussion}


The knowledge in Wikidata is relatively reliable in comparison to other general-domain KGs~\citep{farber2018linked}.
Yet, our analysis reveals a variety of quality aspects of Wikidata that can be improved going forward.
Based on our findings, we propose several recommended actions to include in the interactive contributing environment of Wikidata. These recommendations are intended to prevent low-quality statements from being added, as fixing them later might take a large number of edits. The recommendations can complement ongoing efforts by the Wikidata community to improve data quality based on games and suggestions, aiming to make it easier for users and editors to find and correct mistakes.

\textbf{Integrate entity linking}: To prevent introducing duplicate nodes, it would be beneficial to provide suggestions for similar entities when these exist. For instance, if the user is introducing a basketball player named ``Michael Jordan'' who played for Chicago Bulls, the environment should inform the user that a similar item is already present in Wikidata (with id \texttt{Q41421}).

\textbf{Prevent type and value type violations}: When an editor introduces a new entity, its type should be coherent with the type and value type constraints of its properties. When this is not the case, the editor should be warned about a possible violation. Instead of adapting each new statement, the editor may opt to suggest adapting the constraints themselves.

\textbf{Introduce format guidelines for strings}: Our analysis showed that a large portion of the literal updates transform the literal between two semantically equivalent forms. We propose having more precise formatting guidelines for strings, aiming to adopt consistent naming conventions. For instance, a guideline for initials of human names may dictate including a letter and a dot (``Pamela C. Rasmussen'' rather than ``Pamela C Rasmussen'').

\textbf{Complement missing data}: Wikidata's interactive editing environment should propose that the editor makes complete edits, i.e., edits that satisfy the constraints of the affected properties. 
One way to achieve this would be to suggest that the edits satisfy the constraints of types item-requires-statement, symmetric, and inverse, by either adding the full set of statements that satisfy the constraint, or removing the one violating it. 
A complementary idea is to include a link prediction method, like HINGE~\citep{rosso2020beyond} or StarE~\citep{galkin2020message}, in order to suggest missing statements based on probabilistic graph patterns.

\textbf{Fix statements retroactively}:
Given the large number of existing constraint violations, it is important to help the Wikidata community to fix them. One possibility is to leverage Wikidata's Distributed games\footnote{\url{https://wikidata-game.toolforge.org/distributed/\#}} approach and create games to help editors efficiently validate and fix the constraints. A good starting point for this are the property constraints with large violation ratios, which were detected through our analysis in Table~\ref{tab:vexamples} and Figure~\ref{fig:vrs}. 
An alternative approach, based on our finding in Section~\ref{ssec:complete}, is to fix violations automatically with the expectation that after the automatic fixes there will be fewer violations, and it would be more efficient to fix the errors introduced by the automatic fixes than the original ones. 
Another option is to employ methods that automatically detect errors in KGs~\citep{yao2021typing}.

\section{Related Work}
\label{sec:relatedwork}

The quality of Knowledge Graphs has been studied in existing literature. \cite{chen2019practical} proposed a framework for evaluating the quality of KGs, consisting of dimensions that quantify their fitness for downstream applications. Similarly, quality metrics from 28 prior papers are surveyed by~\cite{piscopo2019we}, and grouped into three dimensions: intrinsic (i.e., accuracy, trustworthiness, and consistency of entities), contextual (i.e., completeness and timeliness of resources), and representation (i.e., understanding, interoperability of entities). 
Our quality indicators are orthogonal to these metrics, as we consider the consensus of the community for them. In addition, our methods go further by proposing an approach to efficiently evaluate some of the metrics proposed by~\cite{piscopo2019we}. 

Many of the metrics proposed by~\cite{piscopo2019we} are covered by~\cite{farber2018linked}, who compare the quality of modern KGs: Wikidata, YAGO, DBpedia, FreeBase, and OpenCyc. \cite{piscopo2018onto} evaluated the quality of Wikidata from an ontological perspective, using indicators related to quantitative measures of classes and instances (e.g., number of instances and number of properties) and of the richness of classes, relations, and properties (e.g., inheritance richness and class hierarchy depth). Prior work has also investigated whether the quality of a knowledge statement in Wikidata depends on the engagement of its editor (leader or contributor)~\citep{Piscopo2017WhatMA,piscopo2018onto}, or the knowledge provenance indicated through the references of a statement~\citep{piscopo2017provenance}.
 Instead, our work performs a systematic analysis of constraint violations, and assesses whether the removal of statements by the community reduces violations. 

Wikidata includes several tools that monitor, analyze, and enforce aspects of quality. The primary sources tool (PST) facilitates a curation workflow for uploading data into Wikidata.\footnote{\url{https://www.wikidata.org/wiki/Wikidata:Primary_sources\_tool\#References}} The Objective Revision Evaluation Service (ORES) 
scores revisions automatically, aiming to detect edits which represent vandalisms.\footnote{\url{https://www.wikidata.org/wiki/Wikidata:ORES}} Recoin (``Relative Completeness Indicator'')~\citep{balaraman2018recoin} extends Wikidata entity pages with information about the relative completeness of the information. Relative completeness is computed by comparing the available information for an entity against other similar entities. Property constraint pages define existing property constraints and report number of violations for a single dump.\footnote{\url{https://www.wikidata.org/wiki/Help:Property\_constraints\_portal}} Our analysis complements the constraint violations reported by Wikidata's pages, by providing in-depth insights about these violations, and abstracting them into findings and recommendations.\footnote{For more information about data quality tools integrated in Wikidata, we refer the reader to: \url{https://www.wikidata.org/wiki/Wikidata:WikiProject\_Data\_Quality} and \url{https://docs.google.com/presentation/d/1rwjqzPaHTsXNNqDc2Op1-qSbcFyaFwOSnkEkStp5L3E/edit}.}

Recently, Wikidata has started moving beyond individual property constraints, representing a higher-level notion of quality in the form of shapes that are meant to provide norms of well-formedness for sub-graphs describing concepts of interest~\citep{thornton2019using}, e.g., human.\footnote{\url{https://www.wikidata.org/wiki/EntitySchema:E10}} These shapes are collected as Schemas.\footnote{\url{https://www.wikidata.org/wiki/Wikidata:WikiProject\_Schemas}} Each schema defines the desired sub-graph topology describing a given concept, using ShEx shape expressions~\citep{thornton2019using}. Schemas are defined through consensus among specific communities (e.g., molecular biology, software engineering, etc.) interested in standardizing concepts relevant to them.\footnote{\url{https://www.wikidata.org/wiki/Wikidata:Database\_reports/EntitySchema\_directory}} We have not addressed the analysis of Wikidata at this level of abstraction, but the approach described in this work can be naturally extended in this direction. 
A similar observation can be made about prior work that encodes Wikidata constraints based on the multi-attributed relational structures (MARS)~\citep{patel2020wikidata}, a formal data model for generalized property graphs devised by~\cite{marx2017logic}.


Recognizing the complexity of the class and type hierarchy in Wikidata, the authors of YAGO4 hand-crafted a new, principled type hierarchy for Wikidata, specifying constraints in SHACL\footnote{\url{https://www.w3.org/TR/shacl/}} 
and OWL~\citep{mcguinness2004owl}; and running scripts to synthesize YAGO by ingesting the data from Wikidata and processing the SHACL expressions. YAGO4 defines constraints on domain and range, disjointness, functionality, and cardinality. The authors report that enforcing these constraints leads to a removal of 132M  statements from Wikidata, i.e., 28\% of all facts. The constraints defined by YAGO4 overlap partially with the constraints in Wikidata studied in this paper. Subsequent work should compare the findings from validating constraints in YAGO4 and Wikidata, and it should generalize the in-depth analysis done in this paper to other KGs like YAGO4.



\cite{rashid2019quality} investigated the evolution of 10 classes from DBpedia over 11 of its releases, measuring aspects of: persistence, consistency, and completeness. This effort resembles our community-based indicator, but it reports analysis over a small data subset, a smaller knowledge graph, and fewer dumps. The goal of \cite{rashid2019quality} is 
to identify potential problems in the data processing
pipeline, which is orthogonal to our goal of detecting low-quality statements in the knowledge graph itself. 

Other work has focused on data validation in KGs. The LOD Laundromat~\citep{beek2014lod} is a large-scale infrastructure that can validate and clean syntactic errors that do not fit the formal specification of RDF, such as bad encoding, undefined URI prefixes, and premature end-of-file markers. \cite{beek2018literally} devise a toolchain for analyzing of the quality of literals in LOD Laundromat's data collection, proposing to automatically improve their value canonization and language tagging.  Our work focuses on errors that cannot be detected by methods that check the syntactic validity of typed literals, like illegal dates, and is thus orthogonal to such prior work. 

Recent work has assessed quality for specific domains. For instance,~\cite{houcemeddine_turki_2020_4445363} report an analysis using ShEx expressions to assess the quality of COVID-19 knowledge in Wikidata. This analysis is more comprehensive than the one reported in our paper, but with a much more limited scope and less generalizable, reflecting the consensus of a specialized community.

Finally, our work relates to efforts that assess the quality of voluntary contributions to large knowledge bases, like Wikipedia~\citep{wilkinson2007cooperation,raman2020classifying} and Open Street Maps~\citep{mooney2012characteristics,fonte2017assessing}. The quality indicators and findings in these works may inspire future research into the quality of large ``wisdom of the crowd''-based KGs like Wikidata.

\section{Conclusions}
\label{sec:conclusion}

This paper studies the quality of Wikidata by proposing three quality indicators based on statements that have been 1) permanently removed; 2) deprecated; or 3) violate constraints defined by the community. Our analysis reveals that, while Wikidata is becoming a KG of increasing quality (removing duplicate entities, fixing modeling errors, and removing constraint violations) there is still room for improvement for preventing entity duplication and constraint violations, having consistent guidelines for literals, and completing missing data. 

Our findings may complement ongoing efforts by the Wikidata community to improve data quality based on games and suggestions, aiming to make it easier for users to find and correct mistakes.
In fact, we are initiating a discussion on how to integrate our methods, findings, and recommendations into Wikidata's infrastructure.
Future work will expand our constraint analysis to additional constraint types and properties; investigate the quality of Wikidata over time, its relation to contributor profiles~\citep{piscopo2018onto}; and will expand our findings by considering additional qualifiers and references~\citep{piscopo2017provenance}. 

\section*{Acknowledgements}

This material is based on research sponsored by Air Force Research Laboratory under agreement number FA8750-20-2-10002. The U.S. Government is authorized to reproduce and distribute reprints for Governmental purposes notwithstanding any copyright notation thereon. The views and conclusions contained herein are those of the authors and should not be interpreted as necessarily representing the official policies or endorsements, either expressed or implied, of Air Force Research Laboratory or the U.S. Government.
Daniel Schwabe was partially supported by grant 309808/2017-0 from CNPq - Brazil.




 \appendix


\bibliographystyle{cas-model2-names}

\bibliography{refs}

\begin{thebibliography}{34}
\expandafter\ifx\csname natexlab\endcsname\relax\def\natexlab#1{#1}\fi
\providecommand{\url}[1]{\texttt{#1}}
\providecommand{\href}[2]{#2}
\providecommand{\path}[1]{#1}
\providecommand{\DOIprefix}{doi:}
\providecommand{\ArXivprefix}{arXiv:}
\providecommand{\URLprefix}{URL: }
\providecommand{\Pubmedprefix}{pmid:}
\providecommand{\doi}[1]{\href{http://dx.doi.org/#1}{\path{#1}}}
\providecommand{\Pubmed}[1]{\href{pmid:#1}{\path{#1}}}
\providecommand{\bibinfo}[2]{#2}
\ifx\xfnm\relax \def\xfnm[#1]{\unskip,\space#1}\fi
\bibitem[{Balaraman et~al.(2018)Balaraman, Razniewski and
  Nutt}]{balaraman2018recoin}
\bibinfo{author}{Balaraman, V.}, \bibinfo{author}{Razniewski, S.},
  \bibinfo{author}{Nutt, W.}, \bibinfo{year}{2018}.
\newblock \bibinfo{title}{Recoin: relative completeness in wikidata}, in:
  \bibinfo{booktitle}{Companion Proceedings of the The Web Conference 2018},
  pp. \bibinfo{pages}{1787--1792}.
\bibitem[{Beek et~al.(2018)Beek, Ilievski, Debattista, Schlobach and
  Wielemaker}]{beek2018literally}
\bibinfo{author}{Beek, W.}, \bibinfo{author}{Ilievski, F.},
  \bibinfo{author}{Debattista, J.}, \bibinfo{author}{Schlobach, S.},
  \bibinfo{author}{Wielemaker, J.}, \bibinfo{year}{2018}.
\newblock \bibinfo{title}{Literally better: Analyzing and improving the quality
  of literals}.
\newblock \bibinfo{journal}{Semantic Web} \bibinfo{volume}{9},
  \bibinfo{pages}{131--150}.
\bibitem[{Beek et~al.(2014)Beek, Rietveld, Bazoobandi, Wielemaker and
  Schlobach}]{beek2014lod}
\bibinfo{author}{Beek, W.}, \bibinfo{author}{Rietveld, L.},
  \bibinfo{author}{Bazoobandi, H.R.}, \bibinfo{author}{Wielemaker, J.},
  \bibinfo{author}{Schlobach, S.}, \bibinfo{year}{2014}.
\newblock \bibinfo{title}{Lod laundromat: a uniform way of publishing other
  people’s dirty data}, in: \bibinfo{booktitle}{International semantic web
  conference}, \bibinfo{organization}{Springer}. pp. \bibinfo{pages}{213--228}.
\bibitem[{Boneva et~al.(2019)Boneva, Dusart, Alvarez and
  Gayo}]{boneva2019shape}
\bibinfo{author}{Boneva, I.}, \bibinfo{author}{Dusart, J.},
  \bibinfo{author}{Alvarez, D.F.}, \bibinfo{author}{Gayo, J.E.L.},
  \bibinfo{year}{2019}.
\newblock \bibinfo{title}{Shape designer for shex and shacl constraints}, in:
  \bibinfo{booktitle}{ISWC 2019-18th International Semantic Web Conference}.
\bibitem[{Chen et~al.(2019)Chen, Cao, Chen and Ding}]{chen2019practical}
\bibinfo{author}{Chen, H.}, \bibinfo{author}{Cao, G.}, \bibinfo{author}{Chen,
  J.}, \bibinfo{author}{Ding, J.}, \bibinfo{year}{2019}.
\newblock \bibinfo{title}{A practical framework for evaluating the quality of
  knowledge graph}, in: \bibinfo{booktitle}{China Conference on Knowledge Graph
  and Semantic Computing}, \bibinfo{organization}{Springer}. pp.
  \bibinfo{pages}{111--122}.
\bibitem[{Dong et~al.(2014)Dong, Gabrilovich, Heitz, Horn, Lao, Murphy,
  Strohmann, Sun and Zhang}]{dong2014knowledge}
\bibinfo{author}{Dong, X.}, \bibinfo{author}{Gabrilovich, E.},
  \bibinfo{author}{Heitz, G.}, \bibinfo{author}{Horn, W.},
  \bibinfo{author}{Lao, N.}, \bibinfo{author}{Murphy, K.},
  \bibinfo{author}{Strohmann, T.}, \bibinfo{author}{Sun, S.},
  \bibinfo{author}{Zhang, W.}, \bibinfo{year}{2014}.
\newblock \bibinfo{title}{Knowledge vault: A web-scale approach to
  probabilistic knowledge fusion}, in: \bibinfo{booktitle}{Proceedings of the
  20th ACM SIGKDD international conference on Knowledge discovery and data
  mining}, pp. \bibinfo{pages}{601--610}.
\bibitem[{F{\"a}rber et~al.(2018)F{\"a}rber, Bartscherer, Menne and
  Rettinger}]{farber2018linked}
\bibinfo{author}{F{\"a}rber, M.}, \bibinfo{author}{Bartscherer, F.},
  \bibinfo{author}{Menne, C.}, \bibinfo{author}{Rettinger, A.},
  \bibinfo{year}{2018}.
\newblock \bibinfo{title}{Linked data quality of dbpedia, freebase, opencyc,
  wikidata, and yago}.
\newblock \bibinfo{journal}{Semantic Web} \bibinfo{volume}{9},
  \bibinfo{pages}{77--129}.
\bibitem[{Fonte et~al.(2017)Fonte, Antoniou, Bastin, Estima, Arsanjani, Bayas,
  See and Vatseva}]{fonte2017assessing}
\bibinfo{author}{Fonte, C.C.}, \bibinfo{author}{Antoniou, V.},
  \bibinfo{author}{Bastin, L.}, \bibinfo{author}{Estima, J.},
  \bibinfo{author}{Arsanjani, J.J.}, \bibinfo{author}{Bayas, J.C.L.},
  \bibinfo{author}{See, L.}, \bibinfo{author}{Vatseva, R.},
  \bibinfo{year}{2017}.
\newblock \bibinfo{title}{Assessing vgi data quality}.
\newblock \bibinfo{journal}{Mapping and the citizen sensor} ,
  \bibinfo{pages}{137--163}.
\bibitem[{Galkin et~al.(2020)Galkin, Trivedi, Maheshwari, Usbeck and
  Lehmann}]{galkin2020message}
\bibinfo{author}{Galkin, M.}, \bibinfo{author}{Trivedi, P.},
  \bibinfo{author}{Maheshwari, G.}, \bibinfo{author}{Usbeck, R.},
  \bibinfo{author}{Lehmann, J.}, \bibinfo{year}{2020}.
\newblock \bibinfo{title}{Message passing for hyper-relational knowledge
  graphs}.
\newblock \bibinfo{journal}{arXiv preprint arXiv:2009.10847} .
\bibitem[{Garijo and Szekely(2021)}]{garijo_daniel_2021_4686805}
\bibinfo{author}{Garijo, D.}, \bibinfo{author}{Szekely, P.},
  \bibinfo{year}{2021}.
\newblock \bibinfo{title}{{Wikidata removed statements from Oct 2014 - Jan
  2021}}.
\newblock \DOIprefix\doi{10.5281/zenodo.4686805}.
\bibitem[{Ilievski et~al.(2020)Ilievski, Garijo, Chalupsky, Divvala, Yao,
  Rogers, Li, Liu, Singh, Schwabe and Szekely}]{ilievski2020kgtk}
\bibinfo{author}{Ilievski, F.}, \bibinfo{author}{Garijo, D.},
  \bibinfo{author}{Chalupsky, H.}, \bibinfo{author}{Divvala, N.T.},
  \bibinfo{author}{Yao, Y.}, \bibinfo{author}{Rogers, C.}, \bibinfo{author}{Li,
  R.}, \bibinfo{author}{Liu, J.}, \bibinfo{author}{Singh, A.},
  \bibinfo{author}{Schwabe, D.}, \bibinfo{author}{Szekely, P.},
  \bibinfo{year}{2020}.
\newblock \bibinfo{title}{Kgtk: a toolkit for large knowledge graph
  manipulation and analysis}, in: \bibinfo{booktitle}{International Semantic
  Web Conference}, \bibinfo{organization}{Springer, Cham}. pp.
  \bibinfo{pages}{278--293}.
\bibitem[{Marx et~al.(2017)Marx, Kr{\"o}tzsch and Thost}]{marx2017logic}
\bibinfo{author}{Marx, M.}, \bibinfo{author}{Kr{\"o}tzsch, M.},
  \bibinfo{author}{Thost, V.}, \bibinfo{year}{2017}.
\newblock \bibinfo{title}{Logic on mars: Ontologies for generalised property
  graphs.}, in: \bibinfo{booktitle}{IJCAI}, pp. \bibinfo{pages}{1188--1194}.
\bibitem[{McGuinness et~al.(2004)McGuinness, Van~Harmelen
  et~al.}]{mcguinness2004owl}
\bibinfo{author}{McGuinness, D.L.}, \bibinfo{author}{Van~Harmelen, F.}, et~al.,
  \bibinfo{year}{2004}.
\newblock \bibinfo{title}{Owl web ontology language overview}.
\newblock \bibinfo{journal}{W3C recommendation} \bibinfo{volume}{10},
  \bibinfo{pages}{2004}.
\bibitem[{M{\"o}ller et~al.()M{\"o}ller, Lehmann and Usbeck}]{mollersurvey}
\bibinfo{author}{M{\"o}ller, C.}, \bibinfo{author}{Lehmann, J.},
  \bibinfo{author}{Usbeck, R.}, .
\newblock \bibinfo{title}{Survey on english entity linking on wikidata} .
\bibitem[{Mooney and Corcoran(2012)}]{mooney2012characteristics}
\bibinfo{author}{Mooney, P.}, \bibinfo{author}{Corcoran, P.},
  \bibinfo{year}{2012}.
\newblock \bibinfo{title}{Characteristics of heavily edited objects in
  openstreetmap}.
\newblock \bibinfo{journal}{Future Internet} \bibinfo{volume}{4},
  \bibinfo{pages}{285--305}.
\bibitem[{Patel-Schneider and Martin(2020)}]{patel2020wikidata}
\bibinfo{author}{Patel-Schneider, P.F.}, \bibinfo{author}{Martin, D.},
  \bibinfo{year}{2020}.
\newblock \bibinfo{title}{Wikidata on mars}.
\newblock \bibinfo{journal}{arXiv preprint arXiv:2008.06599} .
\bibitem[{Piscopo et~al.(2017a)Piscopo, Kaffee, Phethean and
  Simperl}]{piscopo2017provenance}
\bibinfo{author}{Piscopo, A.}, \bibinfo{author}{Kaffee, L.A.},
  \bibinfo{author}{Phethean, C.}, \bibinfo{author}{Simperl, E.},
  \bibinfo{year}{2017}a.
\newblock \bibinfo{title}{Provenance information in a collaborative knowledge
  graph: an evaluation of wikidata external references}, in:
  \bibinfo{booktitle}{International semantic web conference},
  \bibinfo{organization}{Springer}. pp. \bibinfo{pages}{542--558}.
\bibitem[{Piscopo et~al.(2017b)Piscopo, Phethean and
  Simperl}]{Piscopo2017WhatMA}
\bibinfo{author}{Piscopo, A.}, \bibinfo{author}{Phethean, C.},
  \bibinfo{author}{Simperl, E.}, \bibinfo{year}{2017}b.
\newblock \bibinfo{title}{What makes a good collaborative knowledge graph:
  Group composition and quality in wikidata}, in: \bibinfo{booktitle}{SocInfo}.
\bibitem[{Piscopo and Simperl(2018)}]{piscopo2018onto}
\bibinfo{author}{Piscopo, A.}, \bibinfo{author}{Simperl, E.},
  \bibinfo{year}{2018}.
\newblock \bibinfo{title}{Who models the world? collaborative ontology creation
  and user roles in wikidata} \bibinfo{volume}{2}.
\newblock \DOIprefix\doi{10.1145/3274410}.
\bibitem[{Piscopo and Simperl(2019)}]{piscopo2019we}
\bibinfo{author}{Piscopo, A.}, \bibinfo{author}{Simperl, E.},
  \bibinfo{year}{2019}.
\newblock \bibinfo{title}{What we talk about when we talk about wikidata
  quality: a literature survey}, in: \bibinfo{booktitle}{Proceedings of the
  15th International Symposium on Open Collaboration}, pp.
  \bibinfo{pages}{1--11}.
\bibitem[{Raman et~al.(2020)Raman, Sauerberg, Fisher and
  Narayan}]{raman2020classifying}
\bibinfo{author}{Raman, N.}, \bibinfo{author}{Sauerberg, N.},
  \bibinfo{author}{Fisher, J.}, \bibinfo{author}{Narayan, S.},
  \bibinfo{year}{2020}.
\newblock \bibinfo{title}{Classifying wikipedia article quality with revision
  history networks}, in: \bibinfo{booktitle}{Proceedings of the 16th
  International Symposium on Open Collaboration}, pp. \bibinfo{pages}{1--7}.
\bibitem[{Rashid et~al.(2019)Rashid, Torchiano, Rizzo, Mihindukulasooriya and
  Corcho}]{rashid2019quality}
\bibinfo{author}{Rashid, M.}, \bibinfo{author}{Torchiano, M.},
  \bibinfo{author}{Rizzo, G.}, \bibinfo{author}{Mihindukulasooriya, N.},
  \bibinfo{author}{Corcho, O.}, \bibinfo{year}{2019}.
\newblock \bibinfo{title}{A quality assessment approach for evolving knowledge
  bases}.
\newblock \bibinfo{journal}{Semantic Web} \bibinfo{volume}{10},
  \bibinfo{pages}{349--383}.
\bibitem[{Rosso et~al.(2020)Rosso, Yang and
  Cudr{\'e}-Mauroux}]{rosso2020beyond}
\bibinfo{author}{Rosso, P.}, \bibinfo{author}{Yang, D.},
  \bibinfo{author}{Cudr{\'e}-Mauroux, P.}, \bibinfo{year}{2020}.
\newblock \bibinfo{title}{Beyond triplets: hyper-relational knowledge graph
  embedding for link prediction}, in: \bibinfo{booktitle}{Proceedings of The
  Web Conference 2020}, pp. \bibinfo{pages}{1885--1896}.
\bibitem[{Schuler(2005)}]{schuler2005verbnet}
\bibinfo{author}{Schuler, K.K.}, \bibinfo{year}{2005}.
\newblock \bibinfo{title}{Verbnet: A broad-coverage, comprehensive verb
  lexicon} .
\bibitem[{Shenoy et~al.(2021a)Shenoy, Ilievski and
  Garijo}]{shenoy_kartik_2021_5121276}
\bibinfo{author}{Shenoy, K.}, \bibinfo{author}{Ilievski, F.},
  \bibinfo{author}{Garijo, D.}, \bibinfo{year}{2021}a.
\newblock \bibinfo{title}{{Constraint violation summaries (Dump: Dec 7th,
  2020)}}.
\newblock \URLprefix \url{https://doi.org/10.5281/zenodo.5121276},
  \DOIprefix\doi{10.5281/zenodo.5121276}.
\bibitem[{Shenoy et~al.(2021b)Shenoy, Ilievski and
  Garijo}]{kartik_shenoy_2021_5119983}
\bibinfo{author}{Shenoy, K.}, \bibinfo{author}{Ilievski, F.},
  \bibinfo{author}{Garijo, D.}, \bibinfo{year}{2021}b.
\newblock \bibinfo{title}{usc-isi-i2/wd-quality: First notebook release}.
\newblock \URLprefix \url{https://doi.org/10.5281/zenodo.5119983},
  \DOIprefix\doi{10.5281/zenodo.5119983}.
\bibitem[{Shenoy et~al.(2021c)Shenoy, Ilievsky and
  Szekely}]{shenoy_kartik_2021_5120117}
\bibinfo{author}{Shenoy, K.}, \bibinfo{author}{Ilievsky, F.},
  \bibinfo{author}{Szekely, P.}, \bibinfo{year}{2021}c.
\newblock \bibinfo{title}{Wikidata deprecated statements by jan 2021.}
\newblock \URLprefix \url{https://doi.org/10.5281/zenodo.5120117},
  \DOIprefix\doi{10.5281/zenodo.5120117}.
\bibitem[{Surowiecki(2004)}]{surowiecki2004woc}
\bibinfo{author}{Surowiecki, J.}, \bibinfo{year}{2004}.
\newblock \bibinfo{title}{The wisdom of crowds : why the many are smarter than
  the few and how collective wisdom shapes business, economies, societies, and
  nations}.
\newblock \bibinfo{publisher}{Doubleday}, \bibinfo{address}{New York}.
\bibitem[{Tanon et~al.(2020)Tanon, Weikum and Suchanek}]{tanon2020yago}
\bibinfo{author}{Tanon, T.P.}, \bibinfo{author}{Weikum, G.},
  \bibinfo{author}{Suchanek, F.}, \bibinfo{year}{2020}.
\newblock \bibinfo{title}{Yago 4: A reason-able knowledge base}, in:
  \bibinfo{booktitle}{European Semantic Web Conference},
  \bibinfo{organization}{Springer}. pp. \bibinfo{pages}{583--596}.
\bibitem[{Thornton et~al.(2019)Thornton, Solbrig, Stupp, Gayo, Mietchen,
  Prud’Hommeaux and Waagmeester}]{thornton2019using}
\bibinfo{author}{Thornton, K.}, \bibinfo{author}{Solbrig, H.},
  \bibinfo{author}{Stupp, G.S.}, \bibinfo{author}{Gayo, J.E.L.},
  \bibinfo{author}{Mietchen, D.}, \bibinfo{author}{Prud’Hommeaux, E.},
  \bibinfo{author}{Waagmeester, A.}, \bibinfo{year}{2019}.
\newblock \bibinfo{title}{Using shape expressions (shex) to share rdf data
  models and to guide curation with rigorous validation}, in:
  \bibinfo{booktitle}{European Semantic Web Conference},
  \bibinfo{organization}{Springer}. pp. \bibinfo{pages}{606--620}.
\bibitem[{Turki et~al.(2020)Turki, Jemielniak, Taieb, Gayo, Aouicha, Banat,
  Shafee, Prud'Hommeaux, Lubiana, Das and
  Mietchen}]{houcemeddine_turki_2020_4445363}
\bibinfo{author}{Turki, H.}, \bibinfo{author}{Jemielniak, D.},
  \bibinfo{author}{Taieb, M.A.H.}, \bibinfo{author}{Gayo, J.E.L.},
  \bibinfo{author}{Aouicha, M.B.}, \bibinfo{author}{Banat, M.},
  \bibinfo{author}{Shafee, T.}, \bibinfo{author}{Prud'Hommeaux, E.},
  \bibinfo{author}{Lubiana, T.}, \bibinfo{author}{Das, D.},
  \bibinfo{author}{Mietchen, D.}, \bibinfo{year}{2020}.
\newblock \bibinfo{title}{{Using logical constraints to validate information in
  collaborative knowledge graphs: a study of COVID-19 on Wikidata}}.
\newblock \DOIprefix\doi{10.5281/zenodo.4445363}.
\bibitem[{Vrande{\v{c}}i{\'c} and Kr{\"o}tzsch(2014)}]{vrandevcic2014wikidata}
\bibinfo{author}{Vrande{\v{c}}i{\'c}, D.}, \bibinfo{author}{Kr{\"o}tzsch, M.},
  \bibinfo{year}{2014}.
\newblock \bibinfo{title}{Wikidata: a free collaborative knowledgebase}.
\newblock \bibinfo{journal}{Communications of the ACM} \bibinfo{volume}{57},
  \bibinfo{pages}{78--85}.
\bibitem[{Wilkinson and Huberman(2007)}]{wilkinson2007cooperation}
\bibinfo{author}{Wilkinson, D.M.}, \bibinfo{author}{Huberman, B.A.},
  \bibinfo{year}{2007}.
\newblock \bibinfo{title}{Cooperation and quality in wikipedia}, in:
  \bibinfo{booktitle}{Proceedings of the 2007 international symposium on
  Wikis}, pp. \bibinfo{pages}{157--164}.
\bibitem[{Yao and Barbosa(2021)}]{yao2021typing}
\bibinfo{author}{Yao, P.}, \bibinfo{author}{Barbosa, D.}, \bibinfo{year}{2021}.
\newblock \bibinfo{title}{Typing errors in factual knowledge graphs: Severity
  and possible ways out}.
\newblock \bibinfo{journal}{arXiv preprint arXiv:2102.02307} .

\end{thebibliography}

\bio{}
\endbio

\endbio

\section{Sample instantiated query template}\label{ann:queryExample}
The snippet below represents the KGTK queries that encode the item requires statement constraints (IRS) for property P1321 (place of origin (Switzerland)) in Wikidata.\footnote{\url{https://www.wikidata.org/wiki/Property:P1321}} The property has two IRS constraints: 1) each item of the property should be (P31) a human (Q5) and 2) its country of citizenship (P27) should be Switzerland (Q39). There is a single exception to this rule, the person Hans von Flachslanden (Q1583384). The code of the query below is generated automatically with our framework. Comments have been added  (with \emph{"\#"}) to explain the different parts of the query.
 
 \begin{verbatim}
kgtk query  # Query to retrieve valid entities
  -i claims.P1321.tsv  # Statements with property P1321 
     claims.P31.tsv    # Statements with property P31
     claims.P27.tsv    # Statements with property P27 
  --match 
    'P1321: (node1)-[nodeProp]->(node2), 
     P31: (node1)-[]->(node2_P31), 
     P27: (node1)-[]->(node2_P27)' 
  --where 'node2_P31 in ["Q5"]  # subject has to be 
                                # human (Q5)
     and node2_P27 in ["Q39"]'  # subject should live in 
                                # Switzerland (Q39)
  --return 'distinct nodeProp.id, node1 as `node1`, 
    nodeProp.label as `label`, 
    node2 as `node2`'              
  -o claims.P1321.correct_wo_exceptions.tsv             
  --graph-cache cache.db;              

kgtk ifnotexists # Now we calculate violations 
                         # of P1321. 
  -i claims.P1321.tsv             
  --filter-on claims.P1321.correct_wo_exceptions.tsv 
  -o claims.P1321.incorrect_wo_exceptions.tsv

kgtk query # Exclude exceptions, i.e., 
                   # Hans von Flachslanden (Q1583384)
  -i claims.P1321.incorrect_wo_exceptions.tsv                     
  --match 
    '(node1)-[]->()' --where 'node1 in ["Q1583384"]'
  -o claims.P1321.incorrect_w_exceptions.tsv                     
  --graph-cache cache.db;  

kgtk ifnotexists  # Filter exceptions from 
                          # violations file
  -i claims.P1321.incorrect_wo_exceptions.tsv             
  --filter-on claims.P1321.incorrect_w_exceptions.tsv             
  -o claims.P1321.incorrect.tsv;   

kgtk cat # Aggregate correct results.
  -i claims.P1321.correct_wo_exceptions.tsv 
     claims.P1321.incorrect_w_exceptions.tsv                     
  -o claims.P1321.correct.tsv   \end{verbatim}

\end{document}